\documentclass[conference]{IEEEtran}
\usepackage{amsmath,amsfonts,amssymb,amsthm}
\usepackage{times}
\usepackage{comment}
\usepackage{xcolor}
\usepackage{graphicx}
\usepackage{subfigure}
\usepackage{multirow}
\usepackage[numbers]{natbib}
\usepackage{multicol}
\usepackage[bookmarks=true]{hyperref}
\usepackage{makecell} 

\usepackage{floatrow}
\usepackage[linesnumbered,ruled]{algorithm2e}
\usepackage{txfonts}
\newcommand{\bF}{\boldsymbol{F}}
\newcommand{\br}{\boldsymbol{r}}
\newcommand{\bxi}{\boldsymbol{\xi}}
\newcommand{\bA}{\boldsymbol{A}}
\newtheorem{thm}{Theorem}[]

\makeatletter
\newcommand{\printfnsymbol}[1]{%
  \textsuperscript{\@fnsymbol{#1}}%
}
\makeatother

\begin{document}

\title{Geometry of contact: contact planning for multi-legged robots via spin models duality}

\author{Author Names Omitted for Anonymous Review. Paper-ID [127]}

\author{\authorblockN{Baxi Chong\printfnsymbol{1}$^1$,
Di Luo\printfnsymbol{1}$^2$ $^3$ $^4$,
Tianyu Wang$^1$ $^5$, 
Gabriel B. Margolis$^6$, 
Juntao He$^1$ $^5$, \\ 
Pulkit Agrawal$^2$ $^6$,
Marin Solja\v{c}i\'{c}$^7$, and
Daniel I. Goldman$^1$ $^5$}
\authorblockA{$^1$ Department of Physics, Georgia Institute of Technology,
Atlanta, Georgia 30332, USA\\
$^2$NSF AI Institute for Artificial Intelligence and Fundamental Interactions  \\  
$^3$Center for Theoretical Physics, Massachusetts Institute of Technology, Cambridge, MA 02139, USA  \\  
$^4$Department of Physics, Harvard University, Cambridge, MA 02138, USA  \\  
$^5$Institute for Robotics and Intelligent Machines, Georgia Institute of Technology, Atlanta, GA, USA \\  
$^6$ MIT Improbable AI Lab\\  
$^7$ Department of Physics, Massachusetts Institute of Technology, Cambridge, MA 02139, USA
}
}

\maketitle

\begin{abstract}
Contact planning is crucial in locomoting systems. Specifically, appropriate contact planning can enable versatile behaviors (e.g., sidewinding in limbless locomotors) and facilitate speed-dependent gait transitions (e.g., walk-trot-gallop in quadrupedal locomotors).  The challenges of contact planning include determining not only the sequence by which contact is made and broken between the locomotor and the environments, but also the sequence of internal shape changes (e.g., body bending and limb shoulder joint oscillation). Most state-of-art contact planning algorithms focused on conventional robots (e.g. biped and quadruped) and conventional tasks (e.g. forward locomotion), and there is a lack of study on general contact planning in multi-legged robots. In this paper, we show that using geometric mechanics framework, we can obtain the global optimal contact sequence given the internal shape changes sequence. Therefore, we simplify the contact planning problem to a graph optimization problem to identify the internal shape changes. Taking advantages of the spatio-temporal symmetry in locomotion, we map the graph optimization problem to special cases of spin models, which allows us to obtain the global optima in polynomial time. We apply our approach to develop new forward and sidewinding behaviors in a hexapod and a 12-legged centipede. We verify our predictions using numerical and robophysical models, and obtain novel and effective locomotion behaviors.
\end{abstract}

\IEEEpeerreviewmaketitle

\footnotetext[1]{\fontsize{8}{8} Equal contribution. Correspondence to {\href{mailto:bchong9@gatech.edu,diluo@mit.edu}{\texttt{bchong9@gatech.edu, diluo@mit.edu}}}}

\section{Introduction}

Contact planning is crucial for both biological and robotic locomotors. In biological systems, animals with various morphology can coordinate their body segments to make and break contact with substrate. If properly synchronized with internal body movement (e.g., limb retraction/protraction), such sequenced substrate contact can generate effective self-propulsion for stable and deliberate locomotion~\cite{hildebrand1965symmetrical,hildebrand1967symmetrical,Kafkafi1998,full1999templates,ijspeert2007swimming,marvi2014sidewinding,astley2015modulation}. 
However, it can be challenging to replicate biological locomotion in robotic counterparts. For example, uncoordinated contact planning can lead to unstable behaviors in both limbless ~\cite{chong2021frequency,marvi2014sidewinding} and legged~\cite{park2001reflex} systems.

\begin{figure}[t]
\centering
\includegraphics[width=1\linewidth]{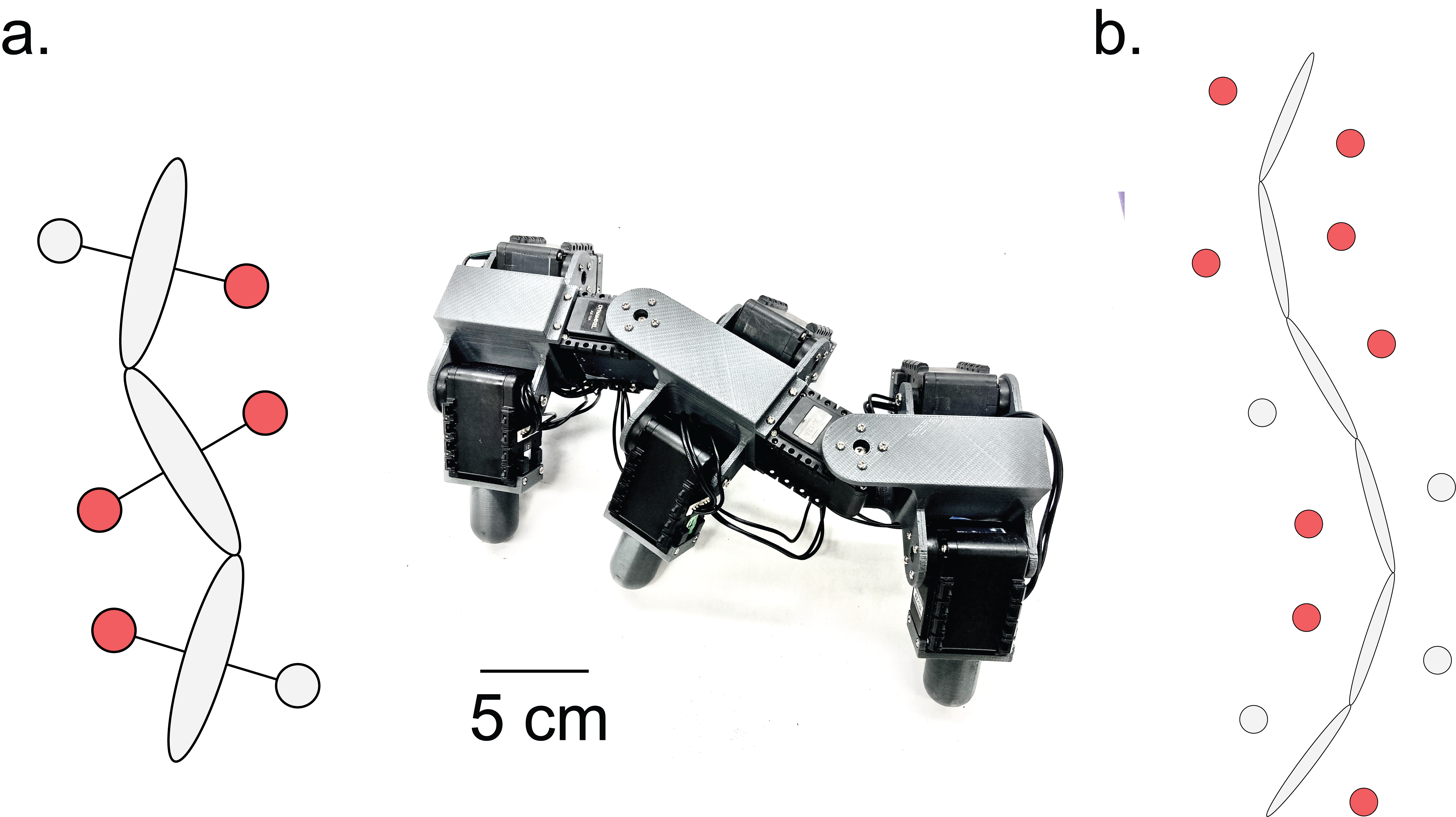}
\caption{\textbf{Multi-legged robot: a geometric and robophysical model} (a) (\textit{Left}) geometric and (right) robophysical hexapod model. (b) Geometric model of a 12-legged centipede.
}
\label{fig:introRobot}
\end{figure}

In practice, contact planning of legged locomotion has been extensively studied. The template-anchor approach~\citet{full1999templates} simplifies the control and contact planning. Specifically, in templates\footnote{``Template" is a behavior that ``contains the smallest number of variables and parameters that exhibits a behavior of interest''~\cite{full1999templates}.}, the complexity of organisms and terrains is ignored with a simple model to reveal general patterns of dynamics. For example, a template for quadrupedal locomotion includes the gait transition from walk to trot and gallop as speed increases~\cite{farley1991mechanical,hildebrand1965symmetrical,hildebrand1967symmetrical}: by prescribing footfall patterns into walk, trot, and gallop, the template simplifies the quadrupedal high-level gait design. When combined with appropriate lower level adaptation to morphological, biomechanical, and environmental details (``anchors'' in the parlance of~\cite{full1999templates}), this approach enables robots with performance approaching those of living systems~\cite{bhounsule2012design,Holmes:2006ku,seok2014design,astley2015modulation}. 

Recent work on some bipedal and quadrupedal robot platforms has demonstrated robust and dynamic locomotion. Most recent research on legged robots focused on the anchor design~\cite{lee2020learning,margolis2022walk,kim2019highly,hwangbo2019learning,fu2021minimizing}, partially because of the well-documented literature on bipedal/quadrupedal templates~\cite{yang2022fast,full1999templates,hildebrand1965symmetrical,hildebrand1967symmetrical}. Recent anchor designs rely on real-time closed-loop feedback control techniques like model-predictive control~\cite{kim2019highly} or reinforcement learning~\cite{hwangbo2019learning}. While effective, these methods demand substantial onboard computation and high-quality proprioceptive or visual sensors.

However, in some applications, sensors can be unreliable and computation scarce. As one example, robots performing search-and-rescue missions must contend with limited vision and possible hardware damage~\cite{whitman2018snake}. Similarly, ultra-cheap and/or ultra-small robots require control methods compatible with their limited sensing and compute. To reduce the cost of building robots and generalize the robotic application, researchers have explored robot locomotion frameworks with less dependence on sensors and controls~\cite{chong2023shannon}. Specifically, prior work has developed hexapod robots~\cite{saranli2001rhex}, on which the additional legs can help avoid catastrophic failures (e.g., loss of stability). Yet, additional legs require  motion pattern coordination and therefore introduce higher dimension for contact planning. To the best of our knowledge, there has been limited research on gait design (contact planning) for hexapod or general multi-legged locomotors. 

In this paper, we establish a geometric mechanics model for general multi-legged locomotors. Specifically, in Sec.~\ref{sec.method}, we establish an elaborated physical model of robot-substrate interactions, and describe the contact planning problem for multi-legged locomotors as a graph optimization problem. Further, using the spatio-temporal symmetry properties in locomotion, we map the graph optimization problem to a special case of a spin model~\cite{mckeehan1925contribution}, which allows us to obtain the global optima in polynomial time. We apply our framework to study hexapod and centipede locomotion. We study the forward and sidewinding behaviors in hexapod locomotion and identify a collection of effective sidewinding gaits for hexapod. We also study centipede (a 12-legged robot) locomotion and obtain effective gaits. We test our hexapod gait predictions using a robophysical model and obtain a good agreement.

\section{Method}\label{sec.method}

\subsection{Geometric mechanics}

\begin{figure}[t]
\centering
\includegraphics[width=1\linewidth]{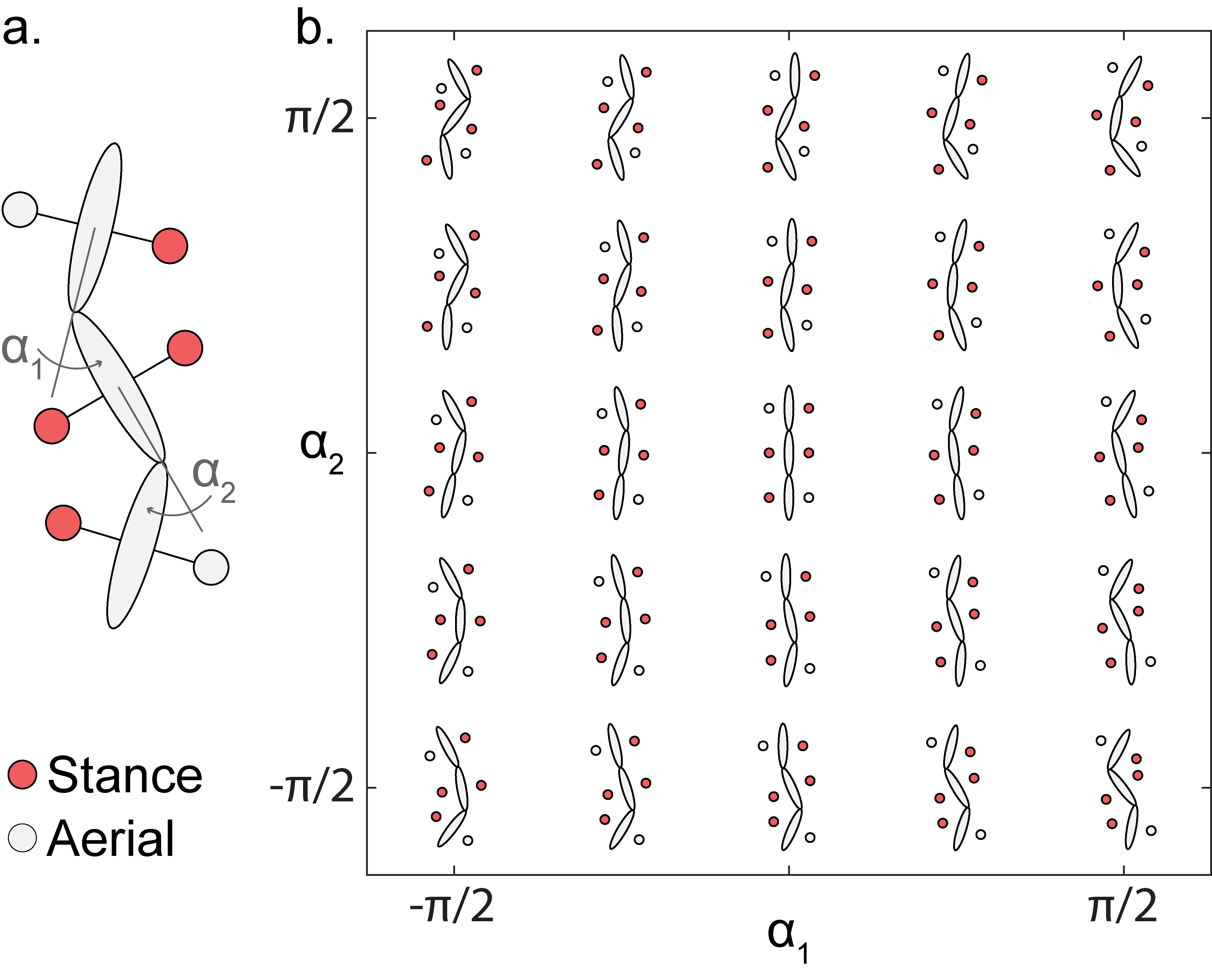}
\caption{\textbf{Shape space of hexapod} (a) Shape variables and contact patterns. Shape variables (body bending joints) are labeled. Limbs in stance and aerial phases are labeled in red and open circles respectively. (b) Shape space of the geometric hexapod model.
}
\label{fig:intro}
\end{figure}

\subsubsection{Kinematic Reconstruction Equation}

In kinematic systems where inertial effects are negligible (compared with frictional forces), the equations of motion \cite{Marsden} can be approximated as

\begin{equation}\label{eq:EquationOfMotion1}
    \bxi={\bA(\br)\dot \br},
\end{equation}
where $\bxi=[\xi_x, \xi_y, \xi_\theta]^T$ denotes the body velocity in the forward, lateral, and rotational directions; ${\br}$ denotes the internal shape variables (joint angles, e.g., $\br=[\alpha_1,\ \alpha_2]$ in Fig.~\ref{fig:intro}); $\bA(\br)$ is the local connection matrix, which encodes environmental constraints and the conservation of momentum. The local connection matrix $\bA$ can be numerically derived using resistive force theory (RFT) to model the ground reaction forces (GRF) \cite{li2013terradynamics,sharpe2015locomotor,zhang2014effectiveness}. Specifically, the net GRF experienced by the locomotor is the sum of the GRF experienced by all segments in stance phase\footnote{We consider a segment to be in stance phase if it is in contact with the ground.}. RFT decomposes the resistive force experienced by a stance-phase segment of a locomotor into two components: $\bF_{\parallel}$ and $\bF_{\perp}$, reaction force along the direction parallel and perpendicular to the body segment respectively. From geometry and physics of GRF, reaction forces of each segment can be calculated from the body velocity $\bxi$, reduced body shape $\br$, and reduced shape velocity $\dot{\br}$~\cite{rieser2019geometric,murray2017mathematical}.
Assuming quasi-static motion, we consider the total net force applied to the system is zero at any instant in time:

\begin{equation}\label{eq:forceIntegral}
    \bF=\sum_{i\in I} {\left[\bF^{i}_{\parallel}\left(\bxi,\br,\dot{\br}\right)+\bF^{i}_{\perp}\left(\bxi,\br,\dot{\br}\right)\right]}=0,
\end{equation}

\noindent where $I$ is the collection of all stance-phase segments. At a given body shape $\br$, Eq.(\ref{eq:forceIntegral}) connects the shape velocity $\dot{\br}$ to the body velocity $\bxi$. Therefore, by the implicit function theorem and the linearization process, we can numerically derive the local connection matrix $\bA(\br)$.
In our implementation, we compute the solution of Eq.(\ref{eq:forceIntegral}) using the MATLAB function \textit{fsolve}. Clearly, the local connection matrix $\bA(\br)$ is dependent on the contact pattern $I$. 

We then consider the relationship between a gait\footnote{``Gait" is a closed-loop path in the shape space.} and its resulting displacement. The displacement along the gait path $ \phi$ can be obtained by integrating the ordinary differential equation (\cite{hatton2015nonconservativity}) below:
\begin{equation}\label{eq:fullapprox}
    g(T) = \int_{ \phi} {L_{g(\br)} \bA(\br) \mathrm{d}{\br}},
\end{equation}

\noindent where $g(\br)= [x(\br),\ y(\br),\ \alpha(\br) ]$ represents the position and rotation of body frame viewed in the world frame at position $r\in \phi$, $T$ is the time period of a gait cycle, and $g(T)=[\Delta x,\ \Delta y,\ \Delta \alpha]$ denotes the translation and rotation of the body frame (w.r.t. the world frame) in one gait cycle. Note that $L_{g}$ is the rotation matrix with respect to $\alpha(\br)$~\cite{murray2017mathematical}.

\begin{figure}[t]
\centering
\includegraphics[width=1\linewidth]{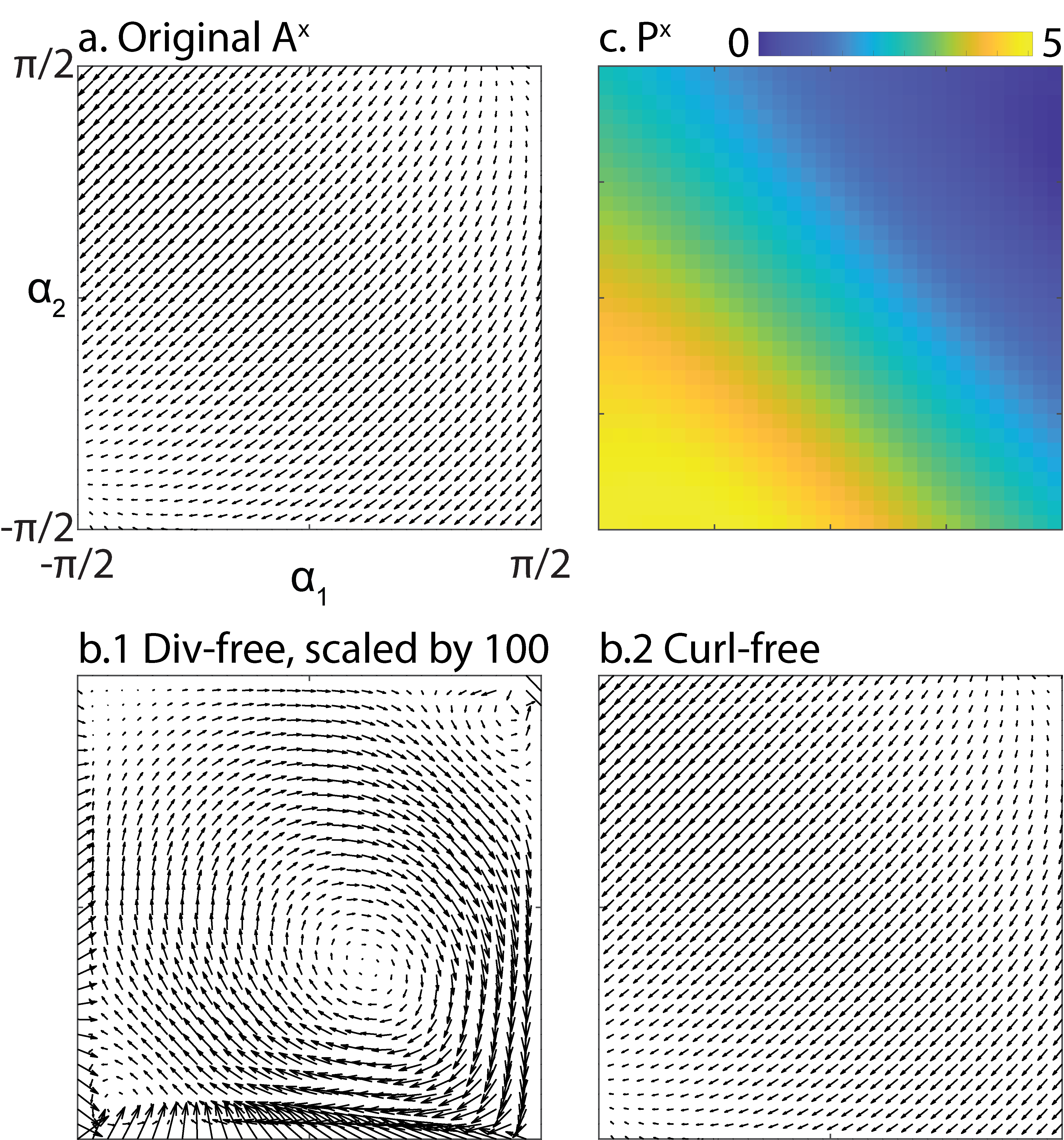}
\caption{\textbf{Connection vector field and Hodge-Helmholtz decomposition} (a) Connection vector field ($\bA^x$) evaluated at contact pattern illustrated in Fig.~\ref{fig:intro}. (b.1) Divergence-free component of the original vector field. Magnitude of the vector field is scaled by 100 for illustration purpose. (b.2) Curl-free component of the original vector field. (c) The potential function computed from the curl-free component of the original vector field. Axes in all panels are identical
}
\label{fig:vec}
\end{figure}

The integral of Eq. (\ref{eq:fullapprox}) can be approximated as:

\begin{equation}\label{eq:lineintegral}
    g(T)
    \approx \int_{ \phi} {\bA(\br)\mathrm{d}\br} =\int_{ \phi}\begin{bmatrix}
\bA^x (\br) \\
\bA^y (\br)\\
\bA^\theta (\br)\\
\end{bmatrix}d\br,
\end{equation}
\noindent where $\bA^x(\br), \bA^y(\br), \bA^\theta(\br)$ are the three rows of the local connections respectively denoting forward, lateral, and rotation vector field. The accuracy of the approximation in Eq.~\ref{eq:lineintegral} can be optimized by properly choosing the body frame \cite{hatton2015nonconservativity,linoptimizing}. An example of vector field ($\bA^x$) is shown in Fig.~\ref{fig:vec}a. With the above derivation, the net displacement over a period can be approximated by the line integral along the gait path over the local connection vector field. 

\subsection{Contact sequence optimization}

Note in prior section, we only consider the locomotion problem with a fixed contact pattern, i.e., $I$ is independent of shape variables. Now we consider the locomotion problem where the contact pattern changes. First, we will explore the contact planning problem with a known shape change sequence. Specifically, let $Q=\big\{r_i,\ i\in\{1,\ 2,\ ...\ M\}\ |\ r_i\in\mathbb{R}^2\big\}$ be a collection of sequenced shape variables. We define that contact switch can only occur in one of the designated shape variables in $Q$. Considering a gait path $\phi$ sequentially connecting all shapes in $Q$, we aim to identify the optimal contact sequence and gait path $\phi$ such that the forward (or lateral) displacement is maximized.

\begin{figure}[t]
\centering
\includegraphics[width=1\linewidth]{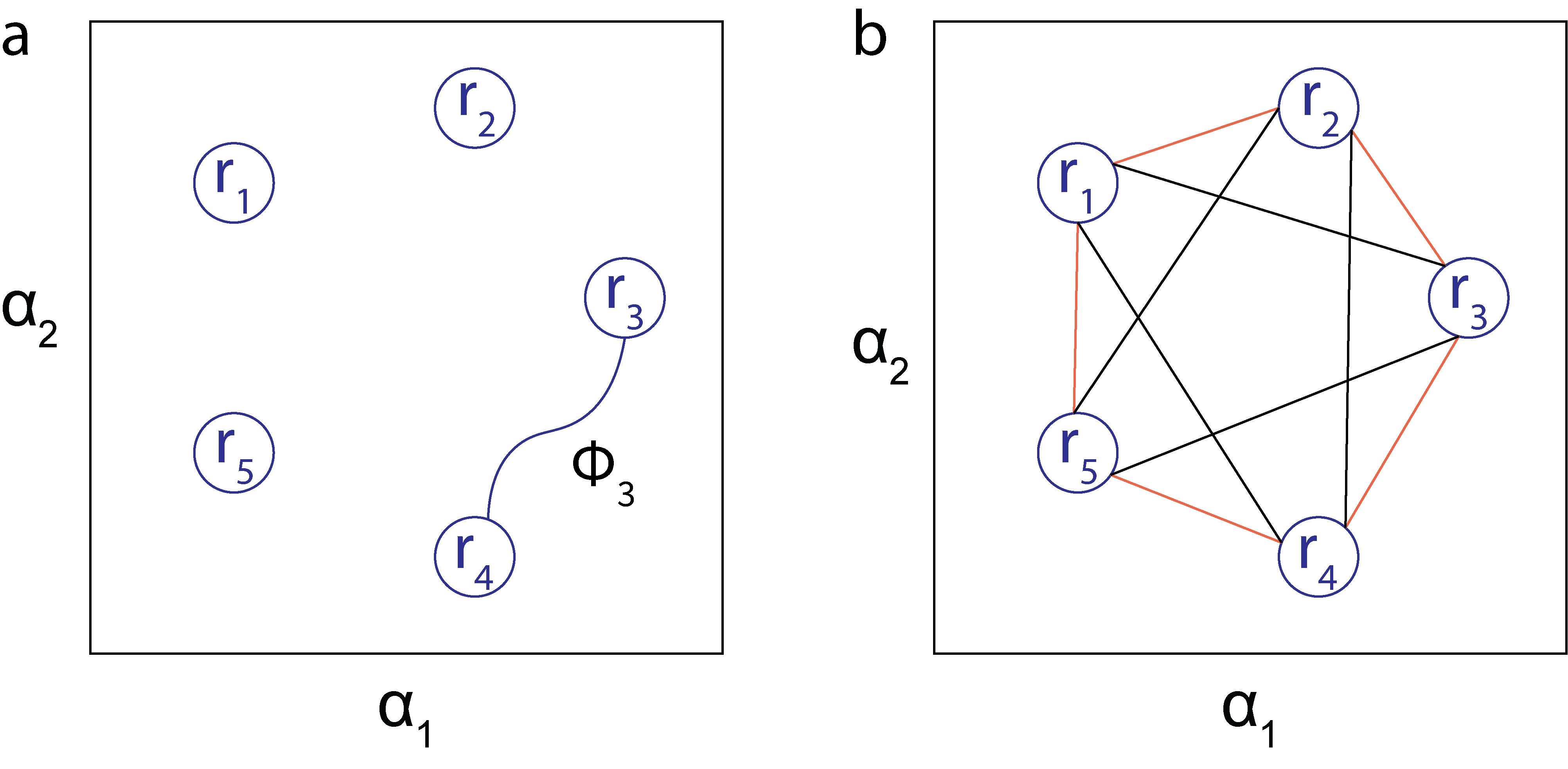}
\caption{\textbf{Gait paths in the shape space} (a) We sample the shape space with a collection of shape variables $\{r_i\}$. $\phi_3$ is an example path connection $r_3$ to $r_4$. $\phi_3$ is a fraction of a close-loop gait path connecting all sampled shape variables. (b) An example of a cheating gait path colored in black. We illustrate the unique non-cheating gait path in red color.
}
\label{fig:cheat}
\end{figure}

We first decompose the closed-loop gait path $\phi$ into piece-wise curves $\phi_i$ connecting $r_i$ to $r_{i+1}$ (Fig.~\ref{fig:cheat}a). The forward displacement from $\phi$ is then:

\begin{equation}\label{eq:piecewiseint}
    \Delta x= \sum_{i=1}^{M} \int_{ \phi_i} {\bA^x_{I(i)}(\br)\mathrm{d}\br},
\end{equation}

\noindent where $I(i)$ is the contact pattern along the piece-wise curves $\phi_i$, and $\bA^x_{I(i)}$ is the forward vector field evaluated at the contact pattern $I(i)$. Note that the optimization in Fig.~\ref{eq:piecewiseint} can be decoupled and separately optimized:

\begin{align}\label{eq:piecewiseint_sep}
    \max_{\phi,\ I(\phi)} \Delta x = & \max_{\phi,\ I(\phi)} \ \sum_{i=1}^{M} \int_{ \phi_i} {\bA^x_{I(i)}(\br)\mathrm{d}\br} \nonumber \\
\end{align}

Now we consider the decoupled optimization problem. Using the Hodge-Helmholtz theorem~\cite{guo2005efficient}, we can decompose a vector field into a curl-free component and a divergence-free component (Fig.~\ref{fig:vec}b). Notably, in most practical locomotion problems on hard ground, the magnitude of divergence-free component is negligible compared with the curl-free component~\cite{chong2021moving,chong2022coordinating,alben2019efficient}, which allows us to focus on the curl-free component of the vector field. For curl-free vector fields, the line-integral is path-independent. Let $P^x_{I(i)}$ be the potential functions of the curl-free components in $\bA^x_{I(i)}$ (Fig.~\ref{fig:vec}c), we can further simplify Eq.~\ref{eq:piecewiseint_sep} to:

\begin{equation}\label{eq:piecewiseint_final}
    \max_{\phi,\ I(\phi)} \Delta x  
     = \max_{\phi,\ I(\phi)} \sum_{i=1}^{M}  \Big( P^x_{I(i)}(r_i) -  P^x_{I(i)}(r_{i+1}) \Big).
\end{equation}

Thus, with the linear and decoupled form, we simplify the contact planning problem to a linear optimization problem in Eq.~\ref{eq:piecewiseint_final}.

\subsection{Shape change sequence optimization}

In the prior section, we assume a pre-fixed shape change sequence and obtain a linear optimization problem formulation for contact planning problem. We then explore the shape change sequence optimization. First, we sample some shapes in the shape space. Let $Q_{all}=\big\{r_i,\ i\in\{1,\ 2,\ ...\ M\}\ |\ r_i\in\mathbb{R}^2\big\}$ be a collection of sampled shape variables. Let $I_{all}$ (with $N$ elements) be a collection of contact states of a multi-legged locomotor. The displacement optimization on $\Delta x(C)$ can be approximated as:

\begin{equation}\label{eq:piecewiseint_final_2}
    \max_{C, \ I(C)} \sum_{i\in C} \Big( P^x_{I(i)}(r_{c(i)}) -  P^x_{I(i)}(r_{c(i+1)}) \Big),
\end{equation}

\noindent where $C\subset M$ is the ordered shape changes, and $\Delta x (C)$ is the resulting displacement. Note that $C$ specifies not only the shapes, but also the sequence of shape changes. In other words, even with the same elements, there could still be different combination of sequences in $C$. However, many of the combination of sequences can be considered as `cheating'. For example, consider a collection of shapes in Fig.~\ref{fig:cheat}b with two combination of sequences colored in red and black. In our framework, we seek to maximize the displacement within \emph{one} period. Therefore, the black path in Fig.~\ref{fig:cheat}b has an unfair advantage over the red path in Fig.~\ref{fig:cheat}b, because the black path winds around the origin twice in one period. To avoid `cheating' and simplify our analysis, we define that the shape can only change in one direction (e.g., clockwise). In this way, given elements in $C$, there is only one valid path connecting all elements. 

In practice, it will always cost  finite time and energy to change contact patterns. To count for such cost, we introduce some penalties in contact switching to the cost function. 

\begin{equation}\label{eq:finalprob}
    \max_{C, \ I(C)}\ \bigg(-\lambda S(C)+ \sum_{i\in C}  \Big( P^x_{I(i)}(r_{c(i)}) -  P^x_{I(i)}(r_{c(i+1)}) \Big) \bigg),
\end{equation}

\noindent where $C$ is a cycle on the graph, $S(C)$ is the number of contact switch over the path $C$, and $\lambda$ is some penalty coefficient. Note that given the explicit physical meaning, we posit that such penalty coefficient $\lambda$ could be empirically measured in future work.  Given the discrete nature of Eq.~\ref{eq:finalprob}, it can be formulated as a graph optimization problem. 

We can further formulate the above optimization problem on as a graph optimization problem. Define $V$ as a collection of nodes $v_{ij}$, where the index $j$ denotes the shape and the index $i$ denotes the contact pattern. At each time step, we can either change contact pattern or shape. Thus, an edge exists connecting two vertices $v_{ij}$ and $v_{kl}$ if $i=k$ or $j=l$. Let $d$ be a set of weights on edges. For an edge connecting $v_{ij}$ and $v_{il}$, its weight $d_{ijil} = P^{I(i)}(r_j) -  P^{I(i)}(r_{l})$.  Note that the weight $d_{ijil}$ can be negative. For an edge connecting $v_{ij}$ and $v_{kj}$, its weight $d_{ijkj}$ is 0, i.e., changing the contact pattern at a fixed shape will not cause displacement. With the above notation, the optimization problem in Eq.~\ref{eq:finalprob} can be reformulated as to find a cycle $C$, such that the sum of weights along $C$ adding together with $-\lambda S(C)$ is maximal. 

Because of the path-independence of connection vector field, we identify the following properties of $d_{ijil}$: (1) Anti-symmetry: $d_{ijil}=-d_{ilij}$, and (2) Additive: $d_{ijil}+d_{ilik}=d_{ijik}$.

\begin{figure}[ht]
\centering
\includegraphics[width=1\linewidth]{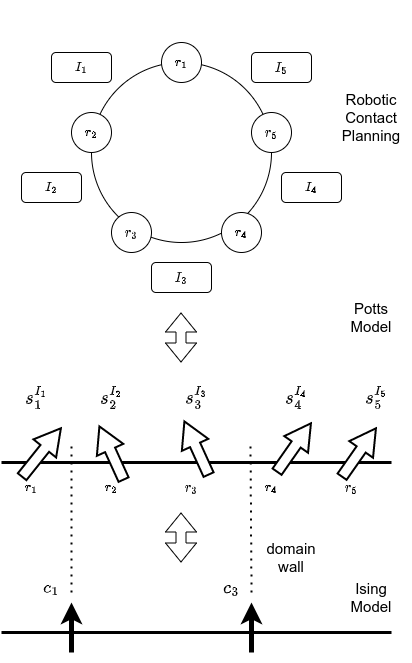}
\caption{\textbf{(Top)} Mapping between robotic contact pattern and Potts model. \textbf{(Bottom)} Duality between Potts model and Ising model.}
\label{fig:mapping}
\end{figure}

\section{Spin models mappings and solutions}

In general, the coupled contact planning and the shape change optimization problem requires search in a space of size $N^M$ where $N$ is the number of contact patterns and $M$ is the number of shape variables. A brute-force search will scale poorly as the number of legs and joints increases. Instead, to obtain a solution efficiently, we develop a connection between the gait optimization problem and two spin models from physics, the Potts model~\cite{potts1952some} and the Ising model~\cite{mckeehan1925contribution}. In this framework, the optimal shape sequence is equivalent to the ground state of the spin models. In this section, we describe the mappings to the Potts model and the Ising Model and how it exploits the special structure of the problem to enable efficient gait optimization.

We start with mapping the shape sequence problem to a $\mathrm{Z}_N$-spin model. A $\mathrm{Z}_N$-spin model considers an array of $M$ spins sitting on $M$ sites, where each spin has $N$ internal states. Each spin can interact with each other or experience an onsite magnetic field. This model is well-known to describe magnetism in physics where each spin can be thought of as a small magnet. In particular, for $N=2$ it reduces to the famous Ising model~\cite{mckeehan1925contribution}. It is known that the Ising models have deep connections to graph optimization problems and provide equivalent formulation of many NP-complete and NP-hard problems~\cite{lucas2014ising}.

\begin{figure}[t!]
\centering
\includegraphics[width=1\linewidth]{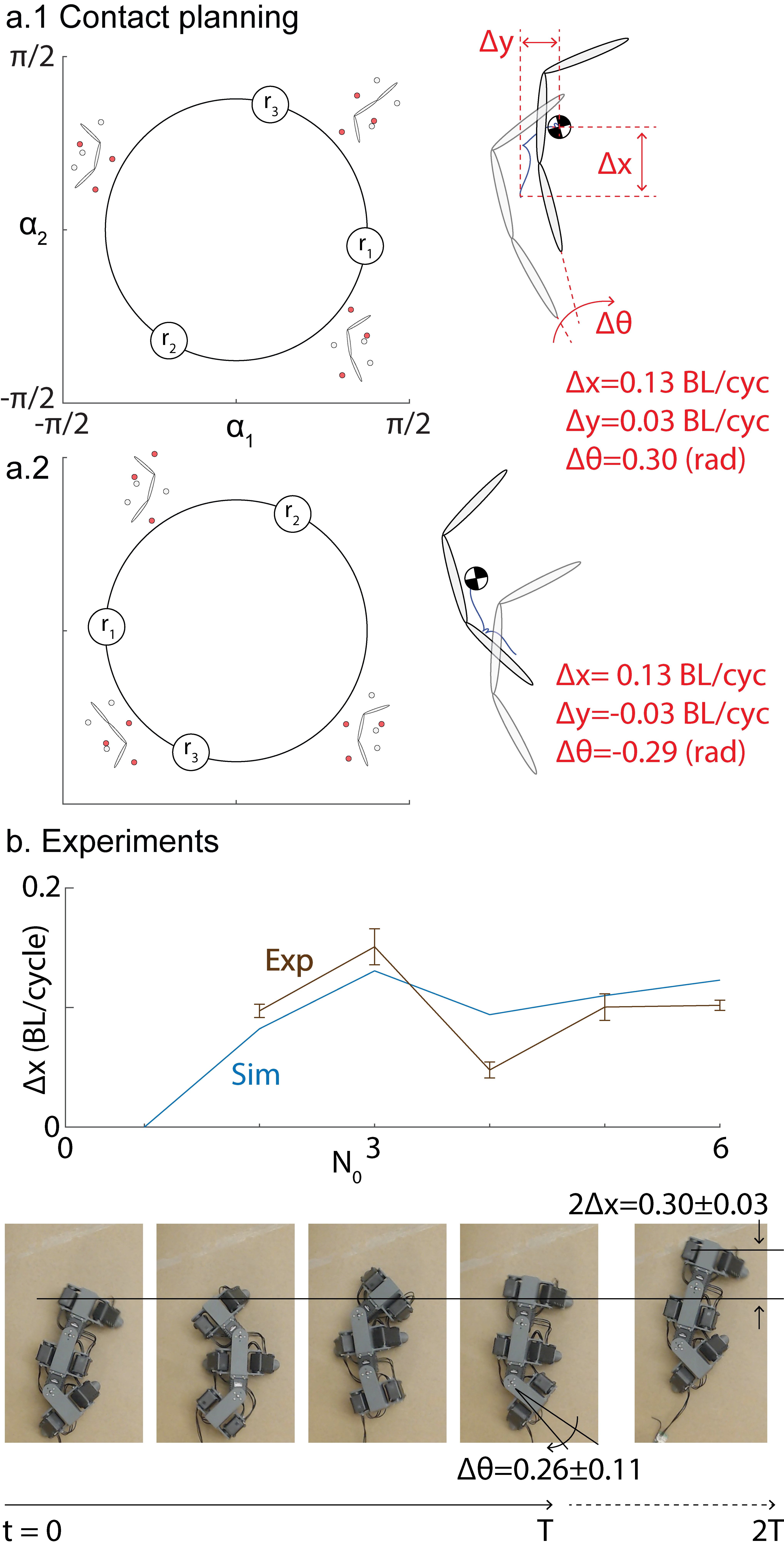}
\caption{\textbf{Contact planning for hexapod forward locomotion}  (a.1) Contact planning for sidewinding hexapod gaits with six contact pattern changes (six-beat) per gait cycle. In the left panel, we illustrate the contact pattern sequence as well as the shape change. In the right panel, we illustrate the net displacement resulting from the contact planning.  Axes in all panels are identical. (a.2) Gait formulation and resulting displacements for the anti-symmetric gaits are illustrated. (b) Numerical and experimental verification. (\textit{Top}) Good agreement between numerical simulation and experiments for sidewinding gaits with different ($N_0$, the number of contact pattern changes) (Bottom) Snapshots of robot implementing six-beat gaits ($N_0 = 6$). 
}
\label{fig:hexforward}
\end{figure}

\subsection{Potts Model}

If we treat the shape variable $r_j$ as the site index $j$ and the contact pattern $I_j$ as the spin internal state, the pair $(r_j,I_j)$ can be mapped to a spin variable $s_j^i$ (or equivalently $s_j^{I_i}$ in Fig.~\ref{fig:mapping}) at site $j$. the displacement $P^x_{I(i)}(r_{c(j)}) -  P^x_{I(i)}(r_{c(j+1)}$ can be associated with the onsite magnetic field energy $d'_{ij(j+1)} s^i_j$ of the spin variable by defining $s^i_j =e^{2\pi i/N}$ and $d'_{ij(j+1)}=d_{ij(j+1)}e^{-2\pi i/N}$. Here $d'_{ijl}$ is interpreted as the onsite magnetic field for the $\mathrm{Z}_N$-spin model. Since the shape sequence optimization problem is defined on a loop, it follows that it can be mapped to a $\mathrm{Z}_N$-spin model with periodic boundary condition as follows:

\begin{equation}
    H = -\sum_j d'_{ij(j+1)} s^i_j
\end{equation}

The optimal shape sequence that maximizes Eq.~\ref{eq:piecewiseint_final_2} is equivalent to the ground state of the Hamiltonian which minimizes the energy. In this case, the ground state is achieved by each individual spin $s^i_j$  follows the largest $d_{ij(j+1)}$. This is also known as the greedy solution. Translating back to the shape sequence language, we have the greedy algorithm as follows.

\textbf{Greedy algorithm.} The solution of optimal displacement in Eq.~\ref{eq:piecewiseint_final_2} is given by

\begin{equation}
      \Delta x (C) =  \sum_{i\in C} \max_{I(i)} \Big( P^x_{I(i)}(r_{c(i)}) -  P^x_{I(i)}(r_{c(i+1)}) \Big),
\end{equation}
For each $r_{c(i)}$ we choose the $I(i)$ that maximizes $P^x_{I(i)}(r_{c(i)}) -  P^x_{I(i)}(r_{c(i+1)}$. \qed

If we further consider the cost of changing contact pattern with penalty $\lambda$ in Eq.~\ref{eq:finalprob}, then it is equivalent to introducing a nearest spin interaction with coupling $\lambda$, which gives rise to a Potts model~\cite{potts1952some} with a nonuniform magnetic field 

\begin{equation}
    H_s = -\sum_j d'_{ij(j+1)} s^i_j - \lambda \delta_{s^i_j, s^i_{j+1}} \label{eq:potts}
\end{equation}
where $\delta_{s^i_j, s^i_{j+1}}=1$ if $s^i_j=s^i_{j+1}$ and 0 otherwise.

The mapping between the shape sequence and the Potts model is also described in the upper part of Fig.~\ref{fig:mapping}. For this model, the search complexity is $N^M$. The ground state is expected to have different phenomena depending on $\lambda$. For small $\lambda$, it is disordered, where each spin follows the local field and the optimal shape sequence is close to the greedy solution. For large $\lambda$, the spins align and it is ferromagnetic. The optimal shape sequence gets close to a uniform gait pattern or gait pattern with only a few jumps.

\begin{figure}[t]
\centering
\includegraphics[width=1\linewidth]{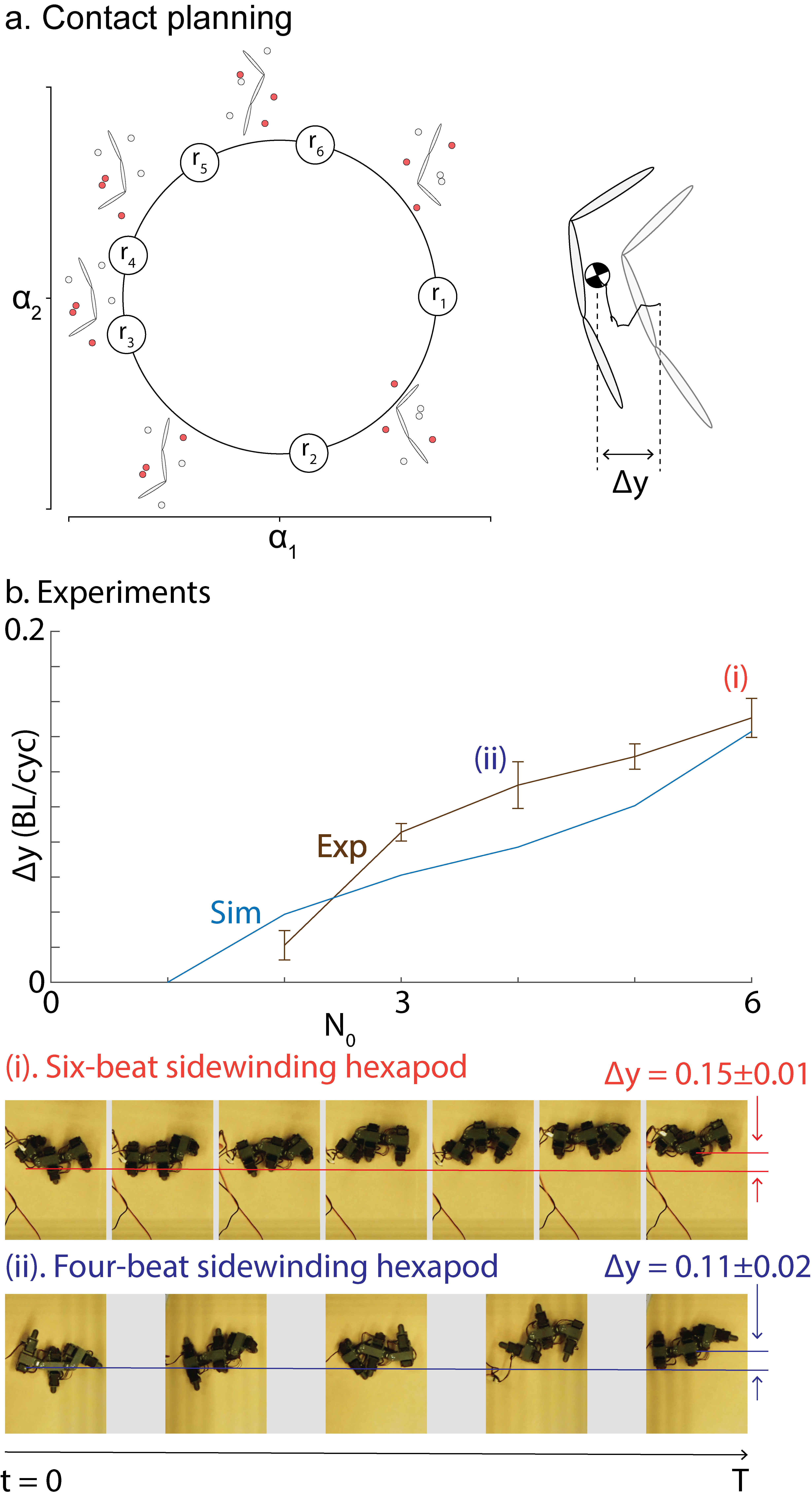}
\caption{\textbf{Contact planning for sidewinding hexapod}  (a) Contact planning for sidewinding hexapod gaits with six contact pattern changes (six-beat) per gait cycle. In the left panel, we illustrate the contact pattern sequence as well as the shape change. In the right panel, we illustrate the net displacement resulting from the contact planning. The axes of shape space is identical to Fig.~\ref{fig:hexforward}a. (b) Numerical and experimental verification. (\textit{Top}) Good agreement between numerical simulation and experiments for sidewinding gaits with different ($N_0$, the number of contact pattern changes) (Bottom) Snapshots of robot implementing six-beat (i, $N_0 = 6$) and four-beat (ii, $N_0 = 6$) gaits. 
}
\label{fig:sidehex}
\end{figure}

\subsection{Ground State Duality to Ising Model}

For the Potts model in Eq.~\ref{eq:potts}, we further define a domain wall $c_{j+1}$ to be the existence of a different spin configuration between $s_j$ and $s_{j+1}$, where $c_{j+1}=1$ if the domain wall exists and otherwise 0. The mapping between the Potts model and the Ising model is shown in the bottom part of Fig.~\ref{fig:mapping}. It follows that we can establish the ground state duality between Potts model and a long-range Ising model using the domain wall mapping and the additive property of $d_{ijl}$. 

\begin{thm}
The Potts model in Eq.~\ref{eq:potts} with $d_{ijl}$ satisfying the additive property has the same ground state energy as the following Ising model with long range couplings $J_{jl}$.

\begin{equation}
    H_c = -\sum_j J_{jl} c_j c_l - \lambda \sum_i c_i 
    \label{eq:domain}
\end{equation}
where $J_{jl} = \textup{max}_i d_{ijl}$.
\end{thm}

\textit{Proof.} Our goal is to show that Eq.~\ref{eq:potts} and Eq.~\ref{eq:domain} admit the same ground state energy. To achieve that, it is sufficient to show each term in Eq.~\ref{eq:potts} can be transformed to a corresponding term in Eq.~\ref{eq:domain}. Consider that there are $K$ locations that the spins do not align with each other in the Potts model, which is equivalent to $K$ domain walls in the Ising model. This costs energy $K\lambda$ from $\lambda s^i_j s^i_{j+1}$ in Eq.~\ref{eq:potts} and correspondingly from $\lambda \sum_i c_i $ in Eq.~\ref{eq:domain}. For every fixed $K$, the minimization of Eq.~\ref{eq:potts} is equivalent to minimize $-\sum_j d'_{ij(j+1)} s^i_j$. This can be achieved by choosing the maximal $\sum_{k=j}^{l-1} d_{ik(k+1)} $ between two domain walls $c_j$ and $c_l$. Due to the additive property of $d_{ijl}$, $\sum_{k=j}^{l-1} d_{ik(k+1)} = d_{ijl}$. It results in the term $-\sum_j J_{jl} c_j c_l$ with $J_{jl} = \textup{max}_i d_{ijl}$ in Eq.~\ref{eq:domain}. \qed

We note that in general Potts model would not be dual to Ising model due to the fact that they have different spin symmetries. However, duality here comes from that we only study the ground state and additive property of $d_{ijl}$ which is special in the robotic contact problem. The mapping reduces the complexity of the original problem from $N^M$ to $2^M$. Based on the Ising model Hamiltonian, we develop a domain-wall search algorithm.

\textbf{Domain-wall search algorithm.} Specify the number of domain-walls or jumps $K$ to be searched. Compute the energy $-\sum J_{ij} c_j c_i$ of all $M$-site spin configurations which have $K$ numbers of $c_j=1$. Choose the lowest energy configuration among all as the solution and translate it back the shape sequence according to the mapping. \qed

There are several advantages of the duality and the domain-wall search algorithm. First, it is independent of the number of shape variables $N$ so that one can always work with a lower degree representation $\mathrm{Z}_2$-spin instead of a $\mathrm{Z}_N$-spin. This is in particularly helpful for generalization to multi-legged robot with many shape variables. Second, the domain wall has a natural interpretation as the jump in the shape sequence problem. In practice, if we would like to have small number of jumps $K$, it is equivalent to impose total spin symmetry number or magnetization $K$ to the model. Without such symmetry the original model in Eq.~\ref{eq:domain} lives in a Hilbert space of size $2^M$ where $M$ is the number of the shape variables, while the symmetry constraint helps to reduce the Hilbert space complexity to $\text{Cr}(M,K)$. We can also ignore the penalty term $\lambda$ in this dual formulation within a fixed spin symmetric sector since the penalty will only contribute a constant shift to the energy in the same sector. It is worth notice that for large number of jumps $K \sim M/2$, $\text{Cr}(M,K) \sim 2^M/\sqrt{M}$ has nearly exponential scaling~\cite{de1730miscellanea}, which opens up opportunities for machine learning applications on dimensional reduction and efficient approximation.

\section{Experimental Setup}
To test our method, we employed a hexapod robot featuring two rotational degrees of freedom in the body and one rotational degree of freedom in each leg. The elongated body, measuring 30 cm in length, is segmented by bending joints that allow for rotations of up to $\pm 90$ degrees. Each segment is equipped with limbs on both sides, which can be lifted and landed through control of the rotational joint at the shoulder. All rotational joints are actuated by Dynamixel AX-12A servo motors, and the body linkages and appendages are 3D-printed using PLA material. During experiments, the body is raised up to 5 cm off the ground, and foot tips were set to be lifted up to 2 cm off the ground. The experiments took place on a flat, hard surface and were based on the assumption of dry Coulomb kinetic friction with a foot-ground friction coefficient $\mu=0.35\pm0.06$. Direct joint angle set-point commands were used to control the robot's movements. Each gait was tested through three repeat experiments, with the robot executing five full cycles per experiment.

The motion of the robot motion was monitored in its position space using an OptiTrack motion capture system comprised of 6 Flex 13 cameras. The 3D positions of IR reflective markers, distributed along the robot body, were recorded at a frame rate of 120 FPS. The data was collected as the robot executed five gait cycles, from the start of the first configuration to the end of the last. The displacement was calculated as the projection of the geometry center trajectory onto the forward/horizontal axes.

\section{Results}

\subsection{Forward hexapod}\label{Sec:forwardHex}

We first apply our framework to study hexapod forward locomotion. We consider stable contact patterns (for hexapod) as at least three legs are in stance phase. We consider the uni-lateral support\footnote{Three legs in stance phase are on the same (either left or right) side.} as unstable contact patterns. Thus, we have 20 contact patterns in total. We sample the shape space by a circular prescription: $\alpha_1 = \pi/3 \sin{(\tau)}, \alpha_2 = \pi/3 \cos{(\tau)}$, and $\tau$ is uniformly sampled over a period.

In robot locomotion, most unstable behaviors in legged locomotion occur during the contact switch~\cite{bai2019cpg,li2016gait}. To avoid excessive contact switches, we explore a hexapod forward gait subject to $N_0$ number of contact pattern changes, and consider the hexapod gaits as a function of $N_0$. Applying our framework, we identify five hexapod forward gaits with $N_0\in[2,\ ...\  6]$. Examples of $N_0= 3$ are illustrated in Fig~\ref{fig:hexforward}a. 

We verify our identified gait using numerical simulation~\cite{chong2022general}. We quantify the displacement in the units of body length travelled per cycle (BL/cyc) We notice that accompanied by the large translation in forward-direction ($\Delta x$), there is non-negligible lateral translation ($\Delta y$) and net rotation ($\Delta \theta$). The emergence of non-negligible lateral translation and rotation can introduce additional challenge to navigate the robot. Thus, we proceed to minimize $\Delta x$ and $\Delta \theta$ associated with the forward locomotion.

We observe that the local connection matrix is symmetric. For any contact pattern $I(i)$, we can obtain its anti-symmetric contact pattern $I'(i)$ such that the contact states of contralateral legs (left and right legs connected to the same body segment) are flipped (Fig.~\ref{fig:hexforward}.a). From symmetry, we know that: $A^x_{I(i)}(\br)=A^x_{I'(i)}(-\br),\ A^y_{I(i)}(\br)=-A^y_{I'(i)}(-\br),\ A^{\theta}_{I(i)}(\br)=-A^{\theta}_{I'(i)}(-\br)$. In other words, with both shape symmetry (e.g., $\br$ and $-\br$) and contact symmetry (e.g., $I(i)$ to $I'(i)$), we can preserve the forward velocity ($\xi_x$) with opposite lateral and rotational velocity ($\xi_y$ and $\xi_\theta$). Thus, we know that for any gait, there exists its anti-symmetric gait with the same forward displacement but opposite lateral displacement and rotation. An example of anti-symmetric gait is shown in the right panel of Fig.~\ref{fig:hexforward}.a. When we execute a gait and its anti-symmetric gait, we can eliminate the unwanted lateral and rotational displacement and obtain forward-only locomotion. 

We test our identified gaits on a robophysical hexapod model. We illustrate that the inclusion of anti-symmetric gaits can eliminate emergence of lateral and rotational displacement. We obtain good agreement between robophysical experiments and numerical simulation (Fig.~\ref{fig:hexforward}.b). Further, we identify hexapod gaits with forward speed up to 0.15 BL/cyc, which we consider as effective given the relatively short legs (leg length $\sim$0.1 BL).

\subsection{Sidewinding hexapod}

Multi-legged locomotors are stereotyped to have better capability to locomote forward instead of sideways, partially because of limb protraction/retraction movements are limited in the fore-aft directions (i.e., no direct contribution from limbs to generate sideway locomotion). Recent work~\cite{chong2021coordination} on quadrupeds revealed that with proper body bending coordination, sideway behavior is possible but with relatively low speed ($<0.1$ BL/cycle). Here, we study the sideway behavior (sidewinding) of hexapod. We posit that with appropriate contact planning, we can enable novel sidewinding behavior of hexapod.

Using the same shape and contact pattern samples from Sec.~\ref{Sec:forwardHex}, we identify sidewinding hexapod gaits as a function of $N_0$. An example of $N_0 = 6$ is shown in Fig.~\ref{fig:sidehex}. We first evaluate the effectiveness of hexapod sidewinding gaits using numerical simulation~\cite{chong2022general}. We notice that hexapod sidewinding gaits have comparable velocity ($0.15\pm0.01$ BL/cyc) as the hexapod forward gaits. 

Finally, we test our identified gaits on robophysical model and obtain good agreement with the numerical simulation. We illustrate the snapshots of robot implementing hexapod sidewinding gaits in Fig.~\ref{fig:sidehex}b.

\begin{figure}[t]
\centering
\includegraphics[width=1\linewidth]{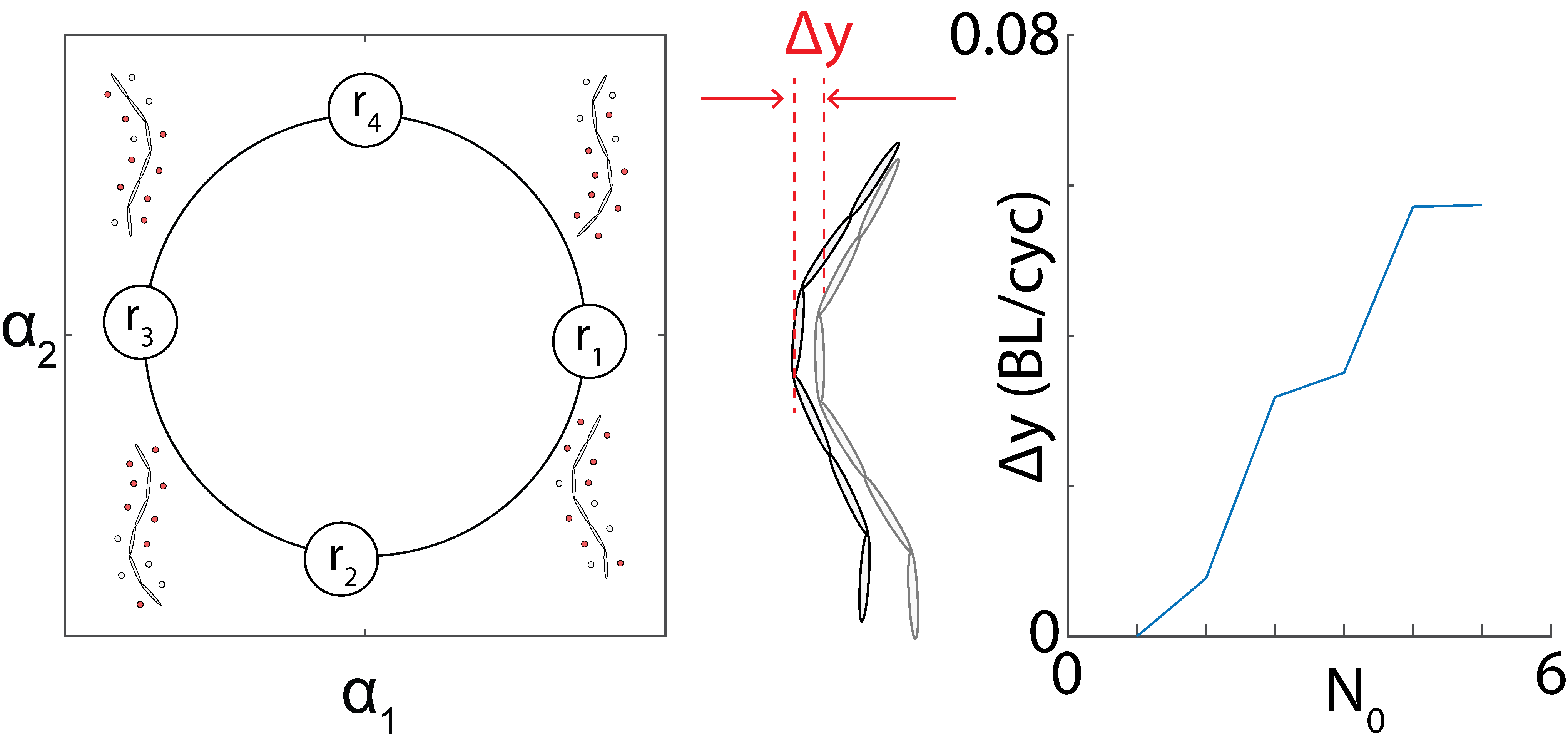}
\caption{\textbf{Contact planning for sidewinding centipede} (\textit{Left}) Contact planning for sidewinding centipede gaits with four contact pattern changes (four-beat) per gait cycle. (\textit{Mid}) The net displacement resulting from the contact planning. (Right) Numerical simulation for sidewinding centipede gaits subject to different $N_0$.
}
\label{fig:centi}
\end{figure}

\subsection{Sidewinding centipede}

Centipede locomotion is challenging partially because of the large number of contact states from the many legs~\cite{chong2022self}. Notably, in our framework, the computational complexity does not scale with the increases in legs. Therefore, we posit that our framework can offer novel insights into centipede locomotion. In this section, we illustrate the generality of our framework by designing sidewinding gaits for a 12-legged centipede locomotor. For simplicity, we consider contact patterns with at least eight legs are in stance phase. Thus, we have 794 contact patterns in total. We sample the shape variable using the shape basis function~\cite{chong2022self}. 

Using our framework, we illustrate effective sidewinding gaits in Fig.~\ref{fig:centi}. From numerical simulation, we notice that the centipede sidewinding gaits have comparable absolute lateral displacement. However, since centipedes have much longer body length, the normalized lateral displacement (BL/cyc) is significantly lower in centipede than hexapod ($0.06 BL/cyc$). With this observation, we hypothesize that the addition in longitudinal length (increasing the leg number) will not improve the locomotors' capability of the sideway locomotion.

\section{Conclusion}

In this paper, we built a geometric mechanics model for general multi-legged locomotion and simplify the contact planning problem to a graph optimization problem. We further develop the mapping between the contact planning problem to both the Potts model and the Ising model, which allows us to solve the optimal shape sequence in polynomial time. We use our framework to study hexapod and centipede (12-legged) locomotion. We obtain effective forward and sidewinding gaits for hexapod in both forward and lateral directions. Further, we identify a collection of gaits as a function of $N_0$, the number of contact changes. We test our prediction on robophysical models. Our work provides a general approach for studying the locomotion of multi-legged robots and the reformulation of the graph optimization problem and spin models open up future opportunities for applications of machine learning. 

In current work, we investigated the locomotion on flat terrains with precise execution of contact patterns. However, in practice, robot locomotion can experience uncertainty in contact patterns. In future work, it is beneficial to establish a probability model on the contact pattern execution related to the finite temperature and entropy properties of the spin models. In doing so, we hypothesize that there exist gaits that are robust over contact uncertainty for reliable locomotion on noisy landscapes. Such robustness then paves ways towards machines that can traverse complex environments with minimal sensing and feedback controls.

\section*{Acknowledgment}
The authors are grateful for funding from NSF-Simons Southeast Center for Mathematics and Biology (Simons Foundation SFARI 594594), Army Research Office (ARO) MURI program, Army Research Office Grant W911NF-11-1-0514 and a Dunn Family Professorship, the NSF AI Institute for Artificial Intelligence and Fundamental Interactions (IAIFI).

\bibliographystyle{plainnat}
\bibliography{references}

\end{document}